# A Modified No Search Algorithm for Fractal Image Compression

Mehdi Salarian, Babak Mohamadinia, Jalil Rasekhi
Electrical Engineering faculty
University Of Mazandaran
Email:mehdi.salarian@gmail.com,

*Abstract*: Fractal image compression has some desirable properties like high quality at high compression ratio, fast decoding, and resolution independence. Therefore it can be used for many applications such as texture mapping and pattern recognition and image watermarking. But it suffers from long encoding time due to its need to find the best match between sub blocks. This time is related to the approach that is used. In this paper we present a fast encoding Algorithm based on no search method. Our goal is that more blocks are covered in initial step of quad tree algorithm. Experimental result has been compared with other new fast fractal coding methods, showing it is better in term of bit rate in same condition while the other parameters are fixed.

## 1.Introduction:

Fractal Image compression is based on the representation of an Image by a set of contractive transforms in different domains [1,10]. This technique uses iterated function system (IFS) theory that has been developed in the past decade. This technique was introduced first by Barnsley [2]. Encoding is not simple, since there is no known algorithm to find the best match and also it requires extensive computations. Although fractal coding suffers from long encoding times, it has advantages such as fast decompression and high compression ratio. Another advantage of fractal image compression is its multi resolution property, i.e. an image can be decoded at higher or lower resolution than the original one, and it is possible to zoom-in on desired sections [3]. These properties caused it be a desirable method for applications in multimedia. The first practical compression scheme was introduced by Jacquin [5] and Jacobs et al [3]. In this scheme an image is partitioned into blocks of size B×B, named range blocks. Also a set of all possible blocks of size 2B×2B, named domain blocks, are constructed. To encode a range block, we must search for an affine transformation and a Domain block so that the transformation maps domain block to range block with minimum error, That means for each rang block, R, we search domain pool to find the suitable domain block D, and the transformation $\tau$ that maps these two blocks, and $\tau(D)$ is the best matching for R. Consider an N×N image, and suppose the size of range block is n×n, Therefore the number of range blocks is $(N/n)^2$, and the number of domain blocks is $(N-2n+1)^2$. The computational cost of finding the best match between a range block and a domain black is $O(n^2)$. If n is constant, the above mentioned complexity will be $O(N^4)$. The problem is that encoding process is very complicated and we need a lot of time to compress images. Many solutions have been proposed for this problem. Almost all of the researchers have presented methods to reduce the search space of the domain pool. Some of them worked on classification of blocks. They considered feature classification or clustering methods [3], which reduce computations by restricting the search process to a subset of domain pool. For example Saupe only used blocks with large variance [6]. This idea is based on the observation that only fractions of domain block with complex structure are used in encoding process. Another algorithm is local search. In this algorithm only domain blocks that surround range block, are selected (fig1.a). But the fastest method that we have seen is no search algorithm that is based on limiting the domain blocks' positions which must be searched [7]. In this paper we present a new approach, namely modified no search (MNS). The main idea is that the best choice for each range block is which, surround it. Fig1.b shows these very well. Encoding is made up from two similar phases; the first is the ordinary search process for finding contrast scaling, S and luminance. Notice that there is only one domain block related each rang block. If the first phase isn't successful, then next phase will be started. In the second phase we try to cover more blocks because the compression rate achieved by a fractal coder is directly related to the number of transforms.

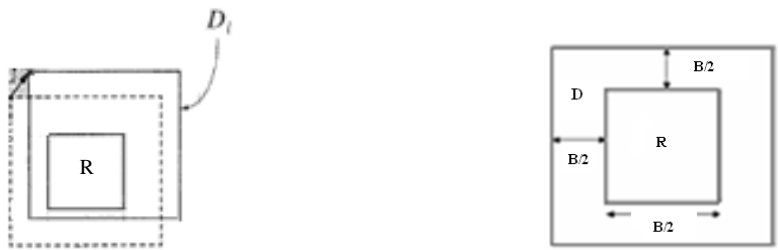

Fig.1. (a): local domain block (b): position relationship between range block and domain block in nosearch

## 2. Local fractal coding

Before anything we would like to know that is there any form of self-similarity help us to decrease encoding time, restricting the domain pool. Many researchers worked in this field. For example Fisher [1] tried to show there isn't any relationship between position of ranges and corresponding domains. He studied on distributions of the difference in the x and y positions of the ranges and selected domain blocks in an exhaustive search. Results shown in figure 2and 3 for lena $512 \times 512$. In this figures, (xr,yr) and (xd,yd) are the range and domain positions. He plotted probability distribution of difference in x and y coordination of two points that have chosen randomly in the unit square with uniform probability, as p(dx)=1-|dx| and p(dy)=1-|dy|. This figures show that even when the points are chosen

Randomly, there is a preference for local domains that means in a full search method, the best domain block for most range blocks are blocks which surrounds them. Therefore we can limit the search area for each rang block. Suppose that image is partitioned in $8 \times 8$ none overlapping blocks and search process for finding suitable block is local and limited to the domain block

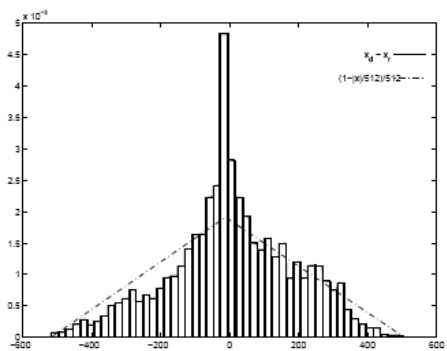

Fig 2 distribution of the difference in the x Position of the domains(xd ) and ranges (xr) for an encoding of $512 \times 512$ lena image

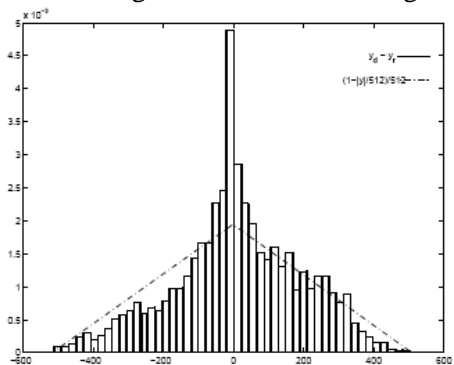

Fig 3 distribution of the difference in the y Position of the domains (yd ) and ranges (yr) for an encoding of $512 \times 512$ lena image.

which surrounds the considered range block. Also no isometric transformation is considered.

Therefore for each range block only 81 blocks would be searched. Here all blocks of size $16 \times 16$, surrounding the range block, are considered.

Implementations show that this method is very fast respect to other methods [6]. S.furao and O.Hasegawa used this idea without mentioning search algorithm. This method will be discussed in section 3.

## 3. No search algorithm

The no search algorithm has been proposed to eliminate the search over domain pool, providing real time applications for the fractal coding. As mentioned, local domain has preference to other strategies. Furthermore statistical results in figure 2, 3 show that the probability of matching is highest when range block located in the center of domain block. Therefore in this method for every range block, the position of corresponding domain block is fixed and only best value for s is searched that causes the encoding be fast enough. To avoid low quality the quadtree scheme is used described below. Some of range blocks mayn't be similar to mentioned domain block and so can't be covered. Therefore they must be covered in a different manner. These range blocks are broken into four equally range blocks, this process can be continued until the range blocks are small enough to be covered with the predefined RMS tolerance. This small range blocks can be covered better than large ones because neighborhood pixels in an image are highly correlated. For implementation, first consider error threshold Ei i =1, 2, 3, 4 for each level. Then partition the image into non-overlapping blocks with size $k \times k$ (suppose center of each block located in (Xi, Yi). For each range block take the domain block with the same center of the range one. Applying mean filter to four neighbor pixels, the size of domain block will be equal to range block. Then find the best value for contrast factor and calculate error between range and domain block. If this error is smaller than corresponding predefine value Ei or the algorithm is in level 4 of quad-tree scheme, we will store the contrast scaling, $s_i$ and luminance, else split the rang block into four equal size blocks and continue search process for them. In above Algorithm four

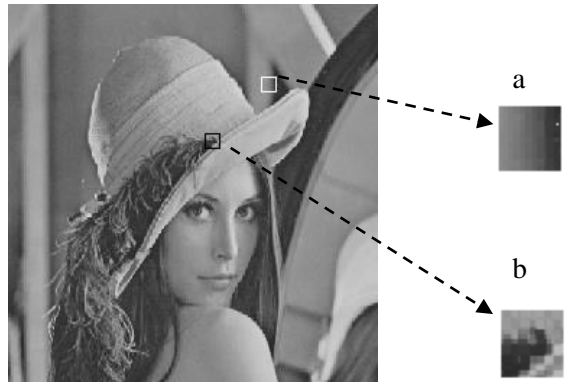

Fig.4: Lena image and two range blocks
a: block with simple structure, b: block with complicated structure levels of quadtree are considered and for each level the Ei values are different. Changing these values, we obtain different results in encoding time, PSNR and compression ratio. This is a drawback of the approach described in section III (no search). We modified no search algorithm by adding a new phase. Here there is only one chance for each range block because we get domain block that its center located in center of it. Then we search for best matching and then calculate error. If error is larger than predefined value of the related level we will split it into four sub blocks then continue search for the new domain block, But in most techniques there is a set of candidate blocks that causes more blocks are covered in higher level and compression ratio is increased. On the other hand, no search algorithm has the lowest encoding time but loses PSNR. To have both speed and quality we propose the below technique. We consider blocks which aren't covered. As shown in figure 4 some of them have complicated structure and must be coded in next levels but in some other it seems that the main reason of fails is difference of luminance in different regions of block. Our goal is to encode this type blocks here efficiently. This procures is described in 3 next subsections.

### 3.1. Technique 1

In this technique like pervious method, for each range block Ri, we calculate mean value Oi. Then we calculate mean value of each sub block. If difference between Oi and all of the Ok k $\in$ {1,2,3,4} is small enough then for each sub block we search best contrast factor from Set S={S1,S2} where S1, S2 are depended to the level the algorithm is. We have four level, therefore we use three set Sj, $j \in \{1,2,3\}$ for levels 1,2,3 respectively. If error values for all sub blocks are desirable, we store related parameters.

## 3.2. Finding the best value for S in phase2

In previous section we explained that we select S from {S1,S2} for each level and these values differ from one level to another that means, we need three set S1, S2, S3. But what is the best choice for S ?. to find the answer we have done many experiments on set of S, results have shown that the best set are S={.2,.5} for level 1 and S={.4,.65} and S={.5,.9} for level 2 and 3 respectively. These sets are the best for all pictures. Here we only use one bit for each sub block.

## 3.3. Storing the parameters

As mentioned before we use four level quadtree structure with the maximum and minimum block size of $16 \times 16$ and $2 \times 2$ respectively to improve the fidelity. As we know the encoding includes two phases. In the first phase for each block we save parameters similar to no-search algorithm. This means that we use one byte for luminance and three bits for contrast. But in second phase we have to store $O_i$, $O_k$ $k \in \{1,2,3\}$ and Sj $j \in \{1,2,3,4\}$. We don't need to store $O_4$ because it calculate simply by

$$1/4\sum_{k=1}^{4} O_k = O_i \text{ then } O_4 = 4O_i - \sum_{k=1}^{3} O_k \quad (1)$$

Related to level number we spend M bits for $O_k$. For example we allocate 5 bits for the first level and 6 bits for other two levels that one of bits determines the values are positive or negative. Addition to this parameter, in all method position of best domain block must be saved. But in no-search method only number of level is saved, that cause to reach to desirable compression ratio. Since we use four level quadtree, we need two bits to address level number for each range block.
This part occupied almost %15 of final compressed image file. Remember that we need extra bit to distinguish between phase1 and phase2. Therefore each range block consumes 14

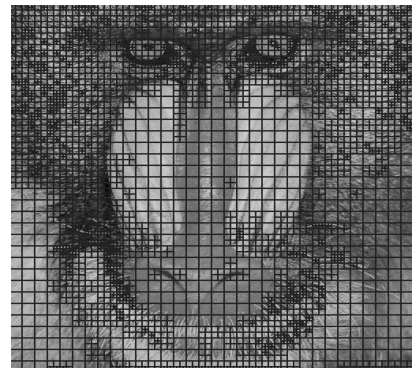
Figure 5.

Bits in phase1. To achieve to better results we use another technique that is described in next section.

## 3.4. Technique 2

Since in most situations specially when original image has complicated structure or when high fidelity is required, a huge number of blocks are encoded in level 4, therefore a lot of memory must be used for level identification. But it isn't necessary to save these two bits for all blocks. Since after encoding of the first block, three next blocks are encoded in this level, therefore it isn't necessary to consume level identification bits for other three blocks. For example consider the Lena image. Suppose that for a reasonable quality 4000 block are covered in level 4. Therefore we need to 8000 identification bits in pervious method, but applying technique2 we need only to 2000 bits. This technique is very significant where the in hand image has complicated structure. For example consider baboon image. As seen in figure 5 almost whole region covered in final level. For example for a medium quality (PSNR=24.5db) almost 25000 block covered in level 4. Applying pervious method 6.25kbytes is needed to encode it. However new method needs only 1.5kbytes. Therefore we can save 4.5 Kbytes.

## 4. The Results

The proposed methods can be evaluated under different points of view. For example a simple distortion metric that is widely used in image compression is the Peak signal to noise (PSNR) which is defined as [4]:

$$PSNR = 10 Log_{10} \frac{k^2}{MSE}$$

$$MSE = \frac{1}{mn}\sum_{i=1}^{m}\sum_{j=1}^{n}(x_{ij} - \hat{x}_{ij})^2$$

Where $x_{ij}$ the pixel is value in original image and $\hat{x}_{ij}$ is the corresponding pixel in the decoded image. K is the maximum intensity value of pixel in image. Since we use 8-bit image, therefore K is 255. The PSNR is measured in dB. This value in fact determines the quality of reconstructed image. Although these distortion metrics are defined without reference to any human perception model, it has been reported that compression system that work as well as for PSNR, will have excellent result in terms of perceptual quality.

the pervious method. For example consider to result of baboon encoding in PSNR equal to 24, in figure6.c. At this point compression ratio for pervious and proposed method are 3.57 and 4.16 respectively. This increasing in compression ratio means that our method needs almost 61.5 Kbytes where no search needs 72 Kbytes. As a result we save 10.5 Kbytes in compressed image file. However this value is very small relative to output file. Therefore distance between two diagrams in figure 6.c is fixed. By Attention to these results we can clearly realize that there are tradeoffs between parameters. When one parameter increases other parameters decrease.

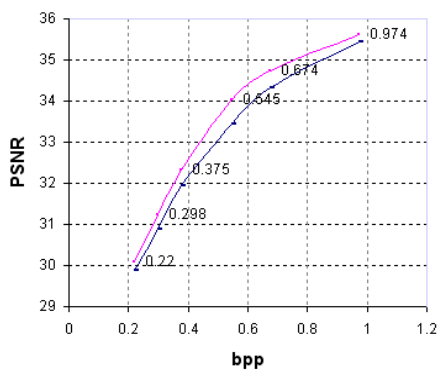
(a)

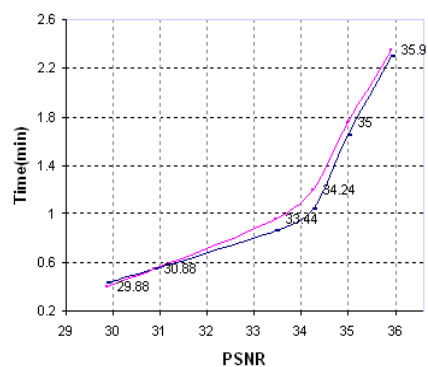
(b)

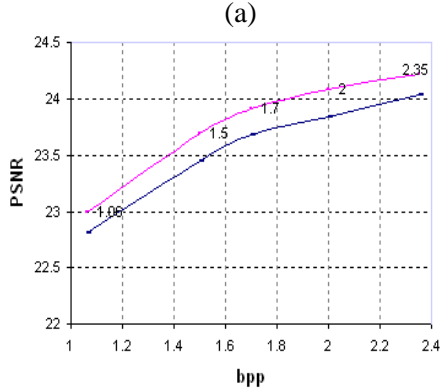
(c)

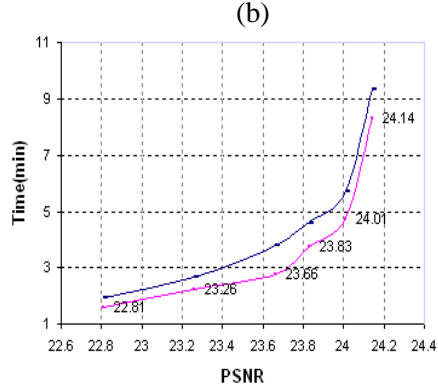
(d)

Figure 6 comparisons between no search and the proposed algorithm.
a, b for lena, c,d for baboon

However there are other parameters such as time and compression ratio that are used by researcher. Result are compared in figures (6) a, b, c and d for no-search and proposed method on Lena and baboon. As you have seen, in the same bit rate our method has better quality. Figure 6.b, 6.d show the diagrams of the encoding time versus bpp. It is obvious that the proposed algorithm is superior to

## 5. Comparison with other method

One of the newest fast techniques that were discussed is adaptive nearest neighbor search by Tong & Wong [9]. This method has compared with no search algorithm in [7,8]. This results show that in the same quality, compression ratio is better in first Method.

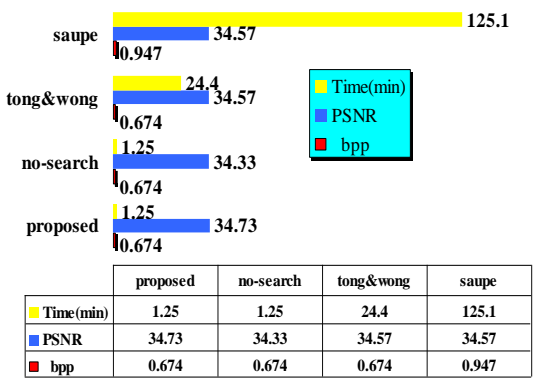

**Figure. 7**

But the encoding time is very better in no-search. For example for Lena image in the same quality no search method is 22 time faster than the method of Tong & Wong. Here not only we decreased encoding time but also we increase the quality of reconstructed image in the same bit rate. Results have been shown in fig.7.

## 6. Conclusions

In this paper, we have examined two last fast fractal image compression algorithms. Then we modified no-search method by using two techniques. In these new methods we tried to cover more blocks in initial level of quadtree scheme. Results show that this method is the fastest method, because it is faster than no-search method. Furthermore experimental result showed that our algorithm is able to achieve a better reconstruction quality in the same compression ratio. In the future we tend to work in wavelet domain to improve the compression ratio.

## Rerfrence: